\documentclass[10pt,twocolumn,letterpaper]{article}

\usepackage[pagenumbers]{cvpr} 

\usepackage{graphicx}
\usepackage{amsmath}
\usepackage{amssymb}
\usepackage{booktabs}
\usepackage{makecell}
\usepackage{multirow}
\usepackage{float}
\usepackage[table]{xcolor}
\usepackage{paralist}

\usepackage[pagebackref=true,colorlinks,bookmarks=false,citecolor=blue,linkcolor=blue,urlcolor=blue]{hyperref}

\usepackage[capitalize]{cleveref}
\usepackage{amsmath,amsfonts,bm}

\crefname{section}{Sec.}{Secs.}
\Crefname{section}{Section}{Sections}
\Crefname{table}{Table}{Tables}
\crefname{table}{Tab.}{Tabs.}

\newcommand{\btheta}{{\boldsymbol{\theta}}}
\newcommand{\bepsilon}{{\boldsymbol{\epsilon}}}

\newcommand{\savespace}{\vspace{-3mm}}
\def\cvprPaperID{241} 
\def\confName{CVPR}
\def\confYear{2023}

\begin{document}

\title{Diffusion-Based Scene Graph to Image Generation with\\ Masked Contrastive Pre-Training}
\author{Ling Yang$^{1*}$\quad Zhilin Huang$^{2}$\thanks{Contributed equally.} \quad Yang Song$^{3}$ \quad Shenda Hong$^{1}$ \quad Guohao Li$^{4}$ \quad Wentao Zhang$^{5}$ \\ Bin Cui$^{1}$ \quad Bernard Ghanem$^{4}$ \quad Ming-Hsuan Yang$^{6,7}$ \\
$^{1}$Peking University \quad $^{2}$Tsinghua University \quad $^{3}$OpenAI \quad $^{4}$KAUST \\ $^{5}$Mila \quad $^{6}$University of California, Merced \quad $^{7}$Google Research\\
{\tt\small yangling0818@163.com, huang-zl22@mails.tsinghua.edu.cn, songyang@openai.com,}\\ {\tt\small wentao.zhang@mila.quebec, \{hongshenda, bin.cui\}@pku.edu.cn,}\\ {\tt\small \{guohao.li, bernard.ghanem\}@kaust.edu.sa, mhyang@ucmerced.edu}
}
\maketitle

\begin{abstract}
Generating images from graph-structured inputs, such as scene graphs, is uniquely challenging due to the difficulty of aligning nodes and connections in graphs with objects and their relations in images. Most existing methods address this challenge by using scene layouts, which are image-like representations of scene graphs designed to capture the coarse structures of scene images. Because scene layouts are manually crafted, the alignment with images may not be fully optimized, causing suboptimal compliance between the generated images and the original scene graphs. To tackle this issue, we propose to learn scene graph embeddings by directly optimizing their alignment with images. Specifically, we pre-train an encoder to extract both global and local information from scene graphs that are predictive of the corresponding images, relying on two loss functions: masked autoencoding loss and contrastive loss. The former trains embeddings by reconstructing randomly masked image regions, while the latter trains embeddings to discriminate between compliant and non-compliant images according to the scene graph. Given these embeddings, we build a latent diffusion model to generate images from scene graphs. The resulting method, called SGDiff, allows for the semantic manipulation of generated images by modifying scene graph nodes and connections. On the Visual Genome and COCO-Stuff datasets, we demonstrate that SGDiff outperforms state-of-the-art methods, as measured by both the Inception Score and Fréchet Inception Distance (FID) metrics. We will release our source code and trained models at \href{https://github.com/YangLing0818/SGDiff}{https://github.com/YangLing0818/SGDiff}.
\end{abstract}

\section{Introduction}
\begin{figure}[t]
\begin{center}\centerline{\includegraphics[width=1\linewidth]{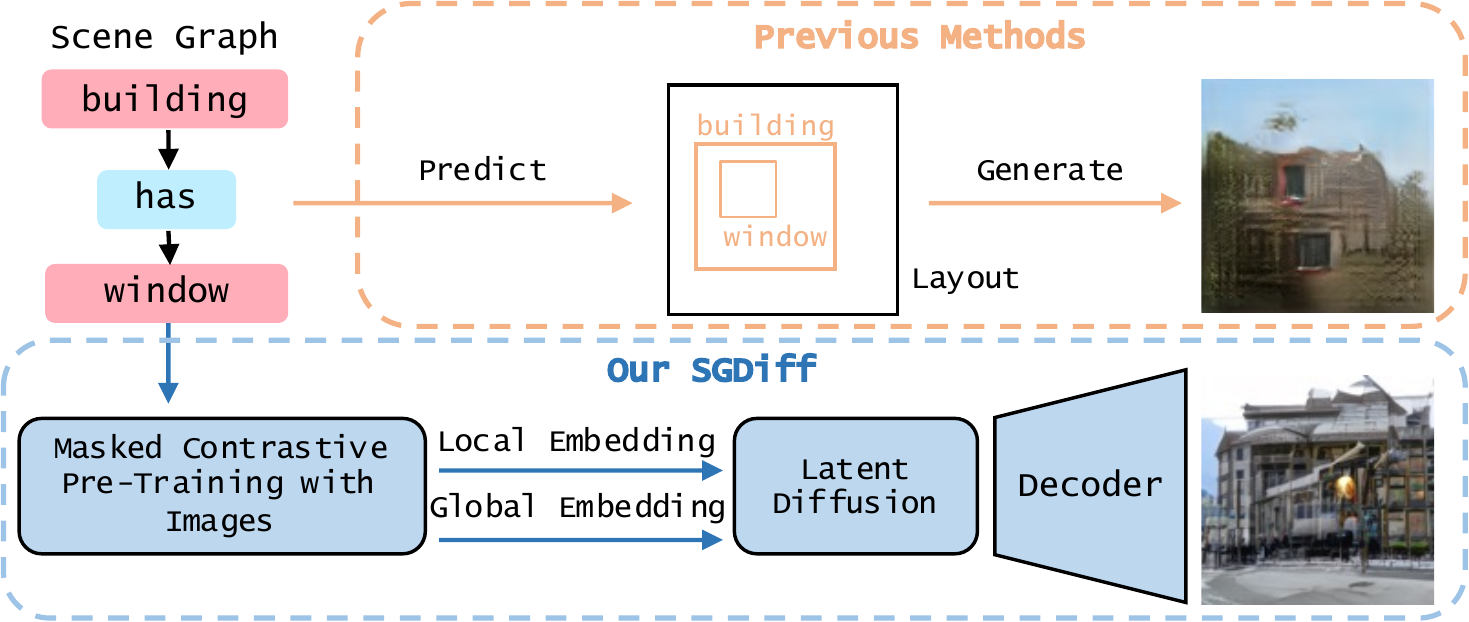}}
\caption{\textbf{SGDiff \vs Previous methods.} Instead of relying on manually specified scene graph representations such as scene layouts, we pre-train our scene graph embeddings with masked autoencoding loss and contrastive learning, explicitly maximizing their predictive power for graph-image alignment. Conditioned on such embeddings, our latent diffusion model \emph{SGDiff} outperforms prior works on scene graph to image generation.}
 
\label{pic-pre}
\end{center}
\vspace{-5mm}
\end{figure}

Image generation has made remarkable progress in the past few years \cite{brock2018large,razavi2019generating,crowson2022vqgan,gafni2022make}, largely due to the success of diffusion and score-based generative models \cite{sohl2015deep,song2019generative,ho2020denoising,song2020score}. These methods allow for the creation of realistic and diverse image samples \cite{dhariwal2021diffusion,ramesh2022hierarchical,saharia2022photorealistic}, which users can specify through various forms---labels \cite{song2020score,dhariwal2021diffusion,ho2022classifier,kim2022diffusionclip}, captions \cite{rombach2022high,saharia2022photorealistic}, segmentation masks \cite{couairon2022diffedit}, sketches \cite{wang2022pretraining}, stroke paintings \cite{meng2021sdedit}, and more \cite{Yang2022DiffusionMA}. However, these types of specifications often fall short when it comes to complex relations between multiple objects in images. Instead, scene graphs provide a concise and accurate way of depicting objects and their relations to one another \cite{johnson2015image,krishna2017visual}. It is therefore crucial to investigate image generation based on scene graphs as a means of synthesizing complex scenes \cite{johnson2018image}.

A key challenge in generating images from scene graphs is ensuring that the resulting image closely aligns with the input scene graph. To this end, generative models must be able to understand the correspondence between the two vastly different data domains: images and graphs. Existing methods \cite{johnson2018image,herzig2020learning,ashual2019specifying,li2019pastegan} mainly address this challenge by using an image-like representation of scene graphs, often in the form of scene layouts, to create coarse sketches for guiding the image generation process. These sketches are then refined by generative models to produce realistic images that follow the specifications given by the scene graph.

While intermediate representations such as scene layouts can be useful, they are often crafted manually and are not specifically designed to facilitate the alignment between images and graphs. For instance, in the case of scene layouts, nodes in scene graphs are usually mapped to bounding boxes and connections are mapped to their spatial layouts. However, not all connections within scene graphs can be accurately translated to spatial layouts, such as \texttt{eating} and \texttt{looking at}. Additionally, some relations, such as \texttt{behind}, \texttt{inside}, and \texttt{in front of}, all correspond to similar spatial relations in scene layouts, creating ambiguity. These intermediate scene layout representations may also contain extraneous information that complicates the training of downstream generative models.

To overcome these limitations, we propose learning intermediate representations that explicitly maximize the alignment between scene graphs and images. We provide an illustration of our approach in \cref{pic-pre}. Specifically, we pre-train a scene graph encoder on graph-image pair datasets to produce embeddings that extract both local and global information from scene graphs, while maximizing their alignment with images. To extract local information, we introduce a masked autoencoding loss, randomly masking out objects in an image and reconstructing the missing portion using unmasked regions and embeddings acquired from the encoder. To gain global information, we leverage contrastive learning to train our encoder to discern between images that do and do not adhere to scene graphs. By combining embeddings obtained from both approaches, we obtain compact intermediate representations of scene graphs that facilitate the alignment between graphs and images.

We showcase the effectiveness of our scene graph embeddings by building a latent diffusion model \cite{vahdat2021score,rombach2022high} that generates images from scene graphs with the aid of our pre-trained embeddings. We evaluate the importance of local and global embeddings through ablation studies, and demonstrate the clear advantages our approach has over traditional intermediate layout representations. Our model, dubbed \emph{SGDiff}, successfully generates images that capture accurate local and global structures of scene graphs. Additionally, our model enables the semantic manipulation of images through scene graph surgery. We evaluate SGDiff on standard datasets such as Visual Genome (VG) \cite{krishna2017visual} and COCO-Stuff \cite{caesar2018coco}, and find that it performs better than current state-of-the-art approaches in both qualitative comparison and quantitative measurements.

\section{Related Work}
\begin{figure*}[ht]
\begin{center}\centerline{\includegraphics[width=1.0\linewidth]{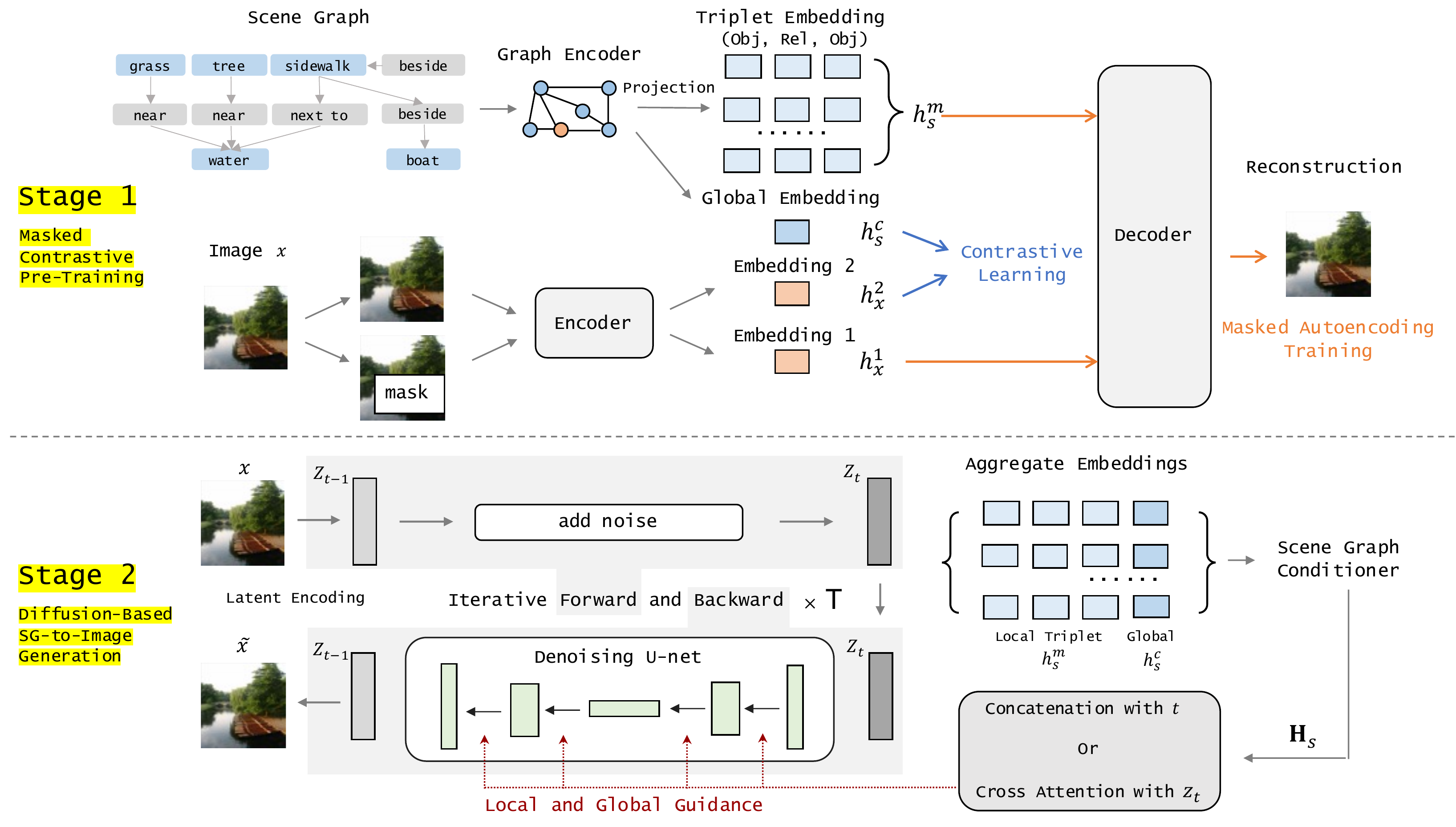}}
\caption{\textbf{Schematic illustration of SGDiff.} In the first stage, we pre-train a scene graph encoder using the combination of a masked autoencoding loss and a contrastive loss. In the second stage, we build a latent diffusion model that conditions on embeddings produced by the scene graph encoder.}
\label{pic-intro}
\end{center}
\end{figure*}
\noindent \textbf{Diffusion Models for Conditional Image Synthesis.}
Diffusion models \cite{Yang2022DiffusionMA} generate data samples by learning to reverse a prescribed diffusion process that converts data to noise. First introduced by Sohl-Dickstein \etal \cite{sohl2015deep} and later improved by Song \& Ermon \cite{song2019generative} and Ho \etal \cite{ho2020denoising}, they are now able to generate image samples with unprecedented quality and diversity \cite{gu2022vector,saharia2022photorealistic,saharia2022palette}.
Diffusion models excel at conditional image synthesis using various forms of user guidance, such as text and images. Existing conditional diffusion models often leverage auxiliary classifiers \cite{song2020score,dhariwal2021diffusion,ho2022classifier,pmlr-v162-nichol22a,kim2022diffusionclip} to incorporate conditional information into the data generating process, using methods of classifier guidance \cite{song2020score,dhariwal2021diffusion} or classifier-free guidance \cite{ho2022classifier}.
Latent Diffusion Models (LDMs) \cite{vahdat2021score,rombach2022high} reduce the training cost for high resolution images by learning the diffusion model in a low-dimensional latent space. They also incorporate conditional information into the sampling process via cross attention \cite{vaswani2017attention}. Similar techniques are employed in DALLE-2~\cite{ramesh2022hierarchical} for image generation from text, where the diffusion model is conditioned on text embeddings obtained from CLIP latent codes \cite{radford2021learning}. Alternatively, Imagen \cite{saharia2022photorealistic} implements text-to-image generation by conditioning on text embeddings acquired from large language models (\eg, T5~\cite{raffel2020exploring}). Despite all this progress on diffusion-based conditional image synthesis, generating images from graph-structured data is under-explored. We fill this gap by designing the first diffusion model for image generation from scene graphs, leveraging new scene graph embeddings constructed from self-supervised learning.

\smallskip
\noindent \textbf{Image Generation from Scene Graphs.}
Scene graphs are graph-structured data for describing multiple objects and their complex relationships in scene images, wherein nodes represent objects and edges represent relations \cite{johnson2015image,krishna2017visual}. Image generation from scene graphs \cite{johnson2018image,tripathi2019using} requires the generative model to reason over both objects and their relations. The first generative model of this kind, Sg2Im \cite{johnson2018image}, proposes a two-stage generation pipeline. First, an embedding model is trained to map scene graphs to scene layouts, which are image-like representations that capture the coarse structure of images to generate. Second, a generative model is trained to refine scene layouts into realistic images. 
Most subsequent work on this task follows the same pipeline. For example, WSGC \cite{herzig2020learning} accounts for semantic equivalence in graph representations by canonicalizing scene graphs before mapping them to scene layouts. Li \etal\cite{li2019pastegan} and Ashual \& Wolf \cite{ashual2019specifying} leverage a repository of external reference images to improve the quality of scene layouts. Other works that rely on scene layout style representations include \cite{zhao2020layout2image,he2021context,sylvain2021object,sun2019image,li2021image}. In lieu of manually crafted scene layouts, we propose learning scene graph embeddings that are both concise and predictive of graph-image alignment. We then use these embeddings to build a latent diffusion model for scene graph to image generation, avoiding the limitations of scene layouts.

\section{Method}
In what follows, we first discuss how to learn effective embeddings of scene graphs via self-supervised learning, then focus on using these embeddings to build diffusion models for scalable scene graph to image generation. 

\subsection{Masked Contrastive Pre-Training}
\label{sec3.1}
Scene layouts are manually constructed representations of scene graphs (SGs). While effective in many cases, they are not specifically optimized to capture all the necessary information from SGs for generating the corresponding scene images. This can lead to suboptimal alignment between SG inputs and generated images. To overcome this limitation, we propose to directly learn such SG representations via self-supervised learning. 
Specifically, given a dataset that contains SG-image pairs, we learn an SG encoder to produce embeddings that capture both local and global information of the input SG, while maximizing their alignment with the corresponding scene images. 
\savespace
\paragraph{Notations.} 
Given a set of objects $\mathcal{C}_o$ and a set of relations $\mathcal{C}_r$, we denote a scene graph $s$ with a tuple $(O, \mathcal{R})$. Here $O= \{o_i \in \mathcal{C}_o\}_{i=1}^n$ represents the set of objects in the scene and $\mathcal{R}=\{r_{ij} \in \mathcal{C}_r\}_{1\leq i,j\leq n}$ denotes the set of relations that exist between these objects. We use a triplet $(o_i, r_{ij}, o_j)$ to denote a directed connection from $o_i$ to $o_j$, representing the relation tuple $(\text{subject}, \text{predicate}, \text{object})$. We denote by $\mathcal{N}_{out}(o_i)$ (resp. $\mathcal{N}_{in}(o_i)$) the set of children (resp. parents) for node $o_i$. To generate an effective representation for $s$, we embed both objects and relations by iterating the following:
\begin{align}
  h_{o_i} &= \text{Pool}(\{f_o^{out}(h_{o_i}, h_{r_{ij}}, h_{o_j})\}_{j \in \mathcal{N}_{out}(o_i)}\notag \\
  &\qquad \cup f_o^{in}(h_{o_j}, h_{r_{ji}}, h_{o_i})\}_{j \in \mathcal{N}_{in}(o_i)}), \\
  h_{r_{ij}} &= f_r(h_{o_j}, h_{r_{ij}}, h_{o_i}),
\end{align}
where $h_{o_i}\in \mathbb{R}^{d_o}$ is the embedding of object $o_j$, $h_{r_{ij}}\in \mathbb{R}^{d_r}$ is the embedding of connection $r_{ij}$ (the relation from $o_i$ to $o_j$), and $f_o^{out}$, $f_o^{in}$, $f_r$ denote separate graph convolutional layers \cite{scarselli2008graph, kipf2017semi}. 
Here $\text{Pool}(\{\})$ denotes the average pooling operator. To train these object and relation embeddings, we oftentimes need an image encoder. The embedding from this encoder is denoted as $h_{x}$ for image $x$.

Below, we introduce two self-supervised techniques for learning object and relation embeddings, focusing on extracting local and global information from SG inputs.

\paragraph{Pretraining with Masked Autoencoding.} Masked pre-training is a preeminent technique in image/text representation learning \cite{devlin2019bert,bao2021beit,he2022masked,chang2022maskgit,xie2022simmim} and visual-language modeling \cite{su2019vl,lu2019vilbert,zhang2021vinvl}. Inspired by its success in these applications, we propose to use a similar technique for learning our SG embeddings.

In particular, we randomly choose a triplet $(o_i, r_{ij}, o_j)$ in the scene graph $s$ and mask out objects $o_i$ and $o_j$ in the corresponding scene image $x$. We denote this masked area as $x_{m}$, and the remaining image as $x_{\backslash m}$. To train the SG encoder, we consider the task of predicting $x_m$ from $x_{\backslash m}$ and embeddings obtained from the SG encoder. Specifically, we concatenate object and relation embeddings from the SG encoder to form $h_s^m$, defined as:
\begin{align}
h_s^m  = \texttt{concat}(\{(f_{o}^m(h_{o_i}),\  f_{r}^m(h_{r_{ij}}),\  f_{o}^m(h_{o_j}))\}), \label{eq:sg}
\end{align}
where $f_{o}^m:\mathbb{R}^{d_o}\rightarrow \mathbb{R}^{d}$, $f_{r}^m:\mathbb{R}^{d_r}\rightarrow \mathbb{R}^{d}$ are MLPs with one hidden layer and ReLU activation functions, and $\texttt{concat}$ stacks all triplets of the form $(f_{o}^m(h_{o_i}),\  f_{r}^m(h_{r_{ij}}),\  f_{o}^m(h_{o_j}))$ in the SG to create the vector $h_s^m$. We jointly train the SG encoder and an auxiliary decoding model $d_\theta$ to minimize the following masked autoencoding loss:
\begin{align}
    \mathcal{L}_{\text{masked}}=\mathop{\mathbb{E}}\limits_{(s, x)\sim D}\Vert x_m - d_\theta(x_{\backslash m}, h_s^m)\Vert_2^2, \label{eq3}
\end{align}
where $(s, x)$ is sampled uniformly at random from $D$, a dataset of graph-image pairs. By training the SG embeddings to reconstruct randomly masked areas in scene images, we explicitly encourage the graph embeddings to encode local structural information that focuses on predicting fine-grained image details from scene graphs.
\savespace
\paragraph{Pretraining with Contrastive Learning.} Contrastive pre-training \cite{chen2020simple,he2020momentum} is a widely adopted technique for learning shared representations across multiple data modalities. They have found success in many visual-language modeling applications \cite{radford2021learning,patashnik2021styleclip}. We propose to leverage contrastive learning as a second way to train our SG embeddings, with a focus on capturing global structural information. Specifically, we compute a graph-level embedding $h_s^c$ by concatenating all object and relation embeddings obtained from the SG encoder, that is,
\begin{align}
    \resizebox{0.85\linewidth}{!}{$\displaystyle h_s^c = f^c(\texttt{concat}(\text{Pool}(\{h_o\}_{o \in \mathcal{O}}),\  \text{Pool}(\{h_r\}_{r \in \mathcal{R}})))$}, \label{eq:sc}
\end{align}
where $f^c:\mathbb{R}^{d_r+d_o}\rightarrow \mathbb{R}^{d}$ is an MLP with one hidden layer and ReLU activation functions. We call $(s, x^+)$ a positive pair if $x^+$ complies with $s$, and $(s, x^-)$ a negative pair if $x^-$ does not adhere to $s$. Positive pairs can be sampled from the graph-image pair dataset $D$, whereas negative pairs are generated by uniformly choosing an image $x^-$ in $D$ that does not match $s$. We optimize the following cross entropy \cite{oord2018representation} loss to learn graph embeddings that can discern positive pairs from negative ones:
\begin{multline}
\mathcal{L}_\text{contrastive}(f; \tau, k) = \\
\resizebox{\linewidth}{!}{$\displaystyle
\mathop{\mathbb{E}}\limits_{\mathop{(s, x^+)\sim D}\limits_{\{x_i^{-}\}_{i=1}^k\mathop{\sim}D_s}}\left[-\log \frac{\exp({{h_s^c}^\top h_{x^{+}} / \tau)}}{ \exp{({h_s^c}^\top h_{x^{+}} / \tau)} + \sum_i\exp{({h_s^c}^\top h_{{x}^{-}_i} / \tau)}}\right]$},
\end{multline}
where $\tau$ is a learnable multiplicative scalar that acts as a temperature parameter, $h_x$ denotes the embeddings of an image $x$ produced by a trainable image encoder, and $D_s$ denotes the set of images in dataset $D$ that do not comply with the SG $s$. This contrastive training objective optimizes our SG embeddings to capture global structures that can identify whether images are in line with the SGs or not.

We combine both the masked autoencoding loss and the contrastive loss to form our training objective for SG embeddings:
\begin{align}
    \mathcal{L} = \mathcal{L}_{\text{masked}} + \lambda  \mathcal{L}_{\text{contrastive}},
\end{align}
where $\lambda > 0$ is a hyperparameter. With this objective function, we can train an effective SG encoder that converts SGs to embeddings without losing predictive information of the matching scene images. Such embeddings provide strong conditioning signals that facilitate downstream diffusion models to generate images from SGs.

\subsection{Diffusion-Based SG to Image Generation}
In diffusion-based conditional image synthesis, we train diffusion models to sample from $p(x \mid y)$, where $x$ is an image, and $y$ is a conditioning signal, typically taking the form of texts in text-to-image generation \cite{saharia2022photorealistic,rombach2022high} and image editing \cite{avrahami2022blended,kawar2022imagic,valevski2022unitune}, or reference images in the context of image translation \cite{meng2021sdedit}. We consider the task of image generation from SGs in this work. With the SG encoder obtained from the previous section, we can build diffusion models to generate images from SGs by setting $y$ to the SG embeddings. For scalable image modeling, we build upon the framework of latent diffusion \cite{vahdat2021score,rombach2022high}, where the diffusion model is trained in a low-dimensional latent space obtained from a pre-trained autoencoder.
\savespace
\paragraph{Diffusion in the Latent Space.} 
We train our diffusion models in a low-dimensional latent space in order to improve the computational efficiency for generating high-resolution images. At the first step, we pre-train an autoencoder to map high-dimensional images to low dimensional latent codes. Specifically, the encoder $f_{enc}$ is trained to transform the image $x\in \mathbb{R}^{H\times W\times 3}$ into a latent code $z\in \mathbb{R}^{h\times w\times c}$ with a downsampling factor $k=H/h=W/w$, and the decoder is trained to reconstruct $x$ from $z$. To regularize the distribution of $z$ for stable generative modeling, we additionally minimize the KL divergence from the distribution of $z$ to a standard Gaussian distribution, in the same vein as Variational Autoencoders \cite{kingma2013auto,rezende2014stochastic}. 

After training both the encoder and the decoder, we use the encoder to generate latent codes for all images in the training dataset, then train a diffusion model on these latent codes separately. In particular, given the latent code $z$ for a randomly sampled training image $x$, we convert it to noise with a Markov process defined by the transition kernel
$q(z_t \mid z_{t-1}) =\mathcal{N}(z_t;\sqrt{\alpha_t}z_{t-1},(1-\alpha_t)\mathbf{I})$, where $t=1, 2, \cdots, T$, 
$z_0 = z$, and $\alpha_{t}$ is a hyper-parameter that controls the rate of noise injection. When the amount of noise is sufficiently large, $z_T$ becomes approximately distributed according to $\mathcal{N}(0, \mathbf{I})$. In order to convert noise back to data for sample generation, we have to estimate the reverse diffusion process by learning the reverse transition kernel $p_{\theta}(z_{t-1} \mid z_t) = \mathcal{N}(\mu_{\theta}(z_t), \Sigma_{\theta}(z_t))$ as an approximation to $q(z_{t-1} \mid z_t)$.
Following Ho \etal \cite{ho2020denoising}, we parameterize $\mu_{\theta}(z_t)$ with a neural network $\bepsilon_{\btheta}(z_{t},t)$ (called the score model \cite{song2019generative,song2020score}) and fix $\Sigma_{\theta}$ to be a constant. The score model can be optimized with denoising score matching \cite{Hyvrinen2005EstimationON,vincent2011connection}. For sample generation, we first generate a latent code $z$ with the diffusion model, then produce an image sample $x$ through the pre-trained decoder.
\savespace
\paragraph{Conditioning on SG Embeddings.}
With the method in \cref{sec3.1}, we can learn an SG encoder to produce two types of scene graph embeddings, $h_s^m$ and $h_s^c$, defined according to \cref{eq:sg} and \cref{eq:sc}. 
To combine these embeddings, we first merge $h_s^c$ and $h_s^m$ by summing $h_s^c$ with each triplet of the form $(f_{o}^m(h_{o_i}),f_{r}^m(h_{r_{ij}}), f_{o}^m(h_{o_j}))$ (\cf, \cref{eq:sg}), which has been generated in the process of computing $h_s^m$. This gives us a new embedding for each relation:
\begin{align}
    h^{sum}_{r_{ij}}= f_{o}^m(h_{o_i}) + f_{r}^m(h_{r_{ij}}) + f_{o}^m(h_{o_j}) + h_s^c.
\end{align}
Afterwards, we concatenate and transform them to get our final embedding, given by 
\begin{align}
\mathbf{H}_s=\psi_{cond}(\texttt{concat}(\{h^{sum}_{r_{ij}}\}_{r_{ij} \in \mathcal{R}})),
\end{align}
where $\psi_{cond}$ is a trainable model called the SG conditioner. We then use the resulting embedding $\mathbf{H}_s$ to guide the generation of our latent diffusion model.

Specifically, we incorporate the embedding into each block of the UNet \cite{ronneberger2015u} backbone of the score model in latent diffusion. We experiment with two different conditioning methods. The first is concatenating the SG embedding $\mathbf{H}_s \in \mathbb{R}^{N \times d_s}$ with time embedding $t$, in which case the conditional score model $\bepsilon_{\btheta}$ takes the form of:
\begin{align}
    \bepsilon_{\btheta}(z_{t},t; \mathbf{H}_s) = \bepsilon_{\btheta}(z_{t},\texttt{concat}(t, \mathbf{H}_s)).
\end{align}
The second is applying cross-attention \cite{vaswani2017attention} to combine the SG embedding $\mathbf{H}_s \in \mathbb{R}^{N \times d_s}$ with the noisy latent code $z_t$, where $\bepsilon_{\btheta}$ is given by
\begin{equation}
    \bepsilon_{\btheta}(z_{t},t; \mathbf{H}_s) = \bepsilon_{\btheta}(\texttt{Cross-Attention}(z_{t},\mathbf{H}_s),t).
\end{equation}
Here $z_t'=\texttt{Cross-Attention}(z_{t},\mathbf{H}_s)$ is defined as:
\begin{equation}
\resizebox{0.9\linewidth}{!}{
 $\displaystyle z_t' =\operatorname{softmax}\left(\frac{(W_Q \cdot  \phi (z_t))(W_K \cdot \mathbf{H}_s)^\top}{\sqrt{d}}\right) \cdot (W_V \cdot \mathbf{H}_s),$
}
\end{equation}
where $W_Q\in \mathbb{R}^{d\times d_{z_t}}$, $W_K\in \mathbb{R}^{d\times d_{\mathbf{H}_s}}$ and $W_V\in \mathbb{R}^{d\times d_{\mathbf{H}_s}}$ are learnable matrices, and $\phi(\cdot)$ is a learnable neural network.
We can directly train these conditional score models to obtain our latent diffusion model, dubbed \emph{SGDiff}.
\section{Experimental Results}
\label{exp}
\begin{figure*}[t]
    \centering
    \includegraphics[width=0.95\textwidth]{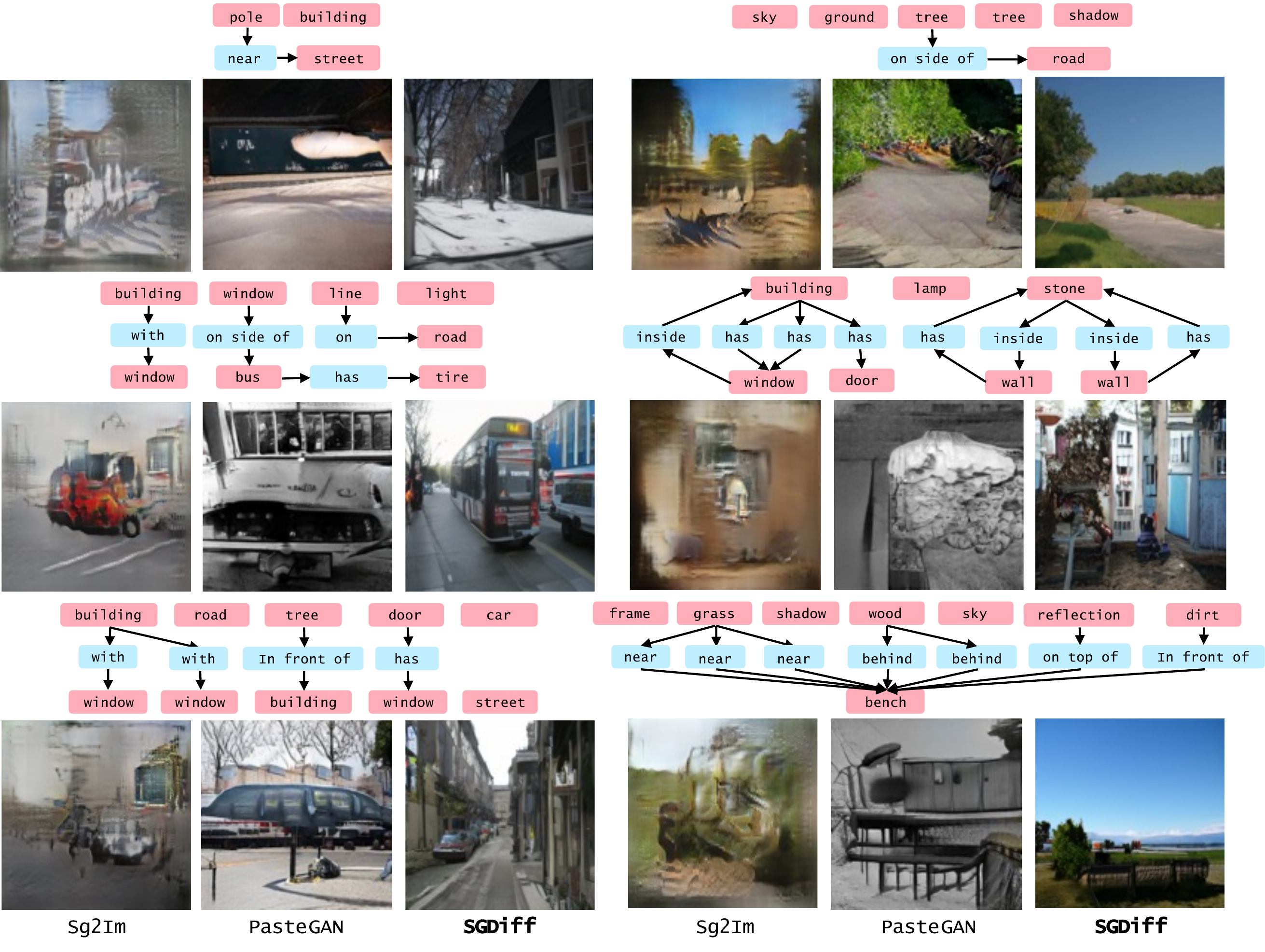}
    \caption{Image samples (128$\times$128) generated by Sg2Im \cite{johnson2018image}, PasteGAN \cite{li2019pastegan}, and our SGDiff given the same scene graphs.}
    \label{exp-qualitative}
    \vskip -0.15in
\end{figure*}
\paragraph{Datasets and Evaluation Metrics.} The Visual Genome (VG) \cite{krishna2017visual} dataset contains 108,077 scene graph \& image pairs, with additional annotations such as bounding boxes and object attributes. Scene graphs in this dataset have 179 object identities, 80 attributes, and 49 relations. For a fair comparison, we follow previous works \cite{johnson2018image,herzig2020learning} to use the standard dataset splits: 62,565 pairs for training, 5,506 for validation, and 5,088 for test. 
COCO-Stuff \cite{caesar2018coco} contains pixel-wise annotations with 40,000 training images and 5,000 validation images with corresponding bounding boxes and segmentation masks. It has 80 item categories and 91 stuff categories. Following \cite{johnson2018image}, we use synthesized scene graphs and standard dataset splits: 25,000 for training, 1,024 for validation, and 2,048 for test. 

We report results with two widely used evaluation metrics. One is Inception Score (IS) \cite{salimans2016improved}, which measures both the quality and diversity of synthesized images. Same as previous work, we employ a pre-trained inception network \cite{szegedy2016rethinking} to obtain network activations for computing IS. The IS is better when larger. The other evaluation metric is Fr\'{e}chet Inception Distance (FID) \cite{heusel2017gans}, which is reported to align well with human evaluation. It measures the distance between the distribution of the generated images and that of the real test images, both modeled as multivariate Gaussians. The FID values are better when lower.

\savespace
\paragraph{Baselines and Implementation Details.} In our experiments, we choose four previous methods on scene graph to image generation as our baselines: Sg2Im \cite{johnson2018image}, WSGC \cite{herzig2020learning}, SOAP \cite{ashual2019specifying}, and PasteGAN \cite{li2019pastegan}. We follow their evaluation settings in all experiments. For masked contrastive pretraining, we train models using the
Adam optimizer \cite{kingma2015adam} with a learning rate of 5e-4 and a batch size of 64 for 100,000 iterations. For latent diffusion training, we train models from scratch using the same optimizer but with a learning rate of 1e-6 and a batch size of 16 for 700,000 iterations. Please consult the \cref{imple-detail} for more details.
\subsection{Quantitative Comparisons}
Similar to prior works \cite{johnson2018image,herzig2020learning,ashual2019specifying,li2019pastegan}, we perform quantitative comparisons on the COCO-Stuff and VG datasets for image resolution 64$\times$64, 128$\times$128, and 256$\times$256. All IS and FID results are provided in \cref{exp-main}. We observe that our SGDiff consistently outperforms existing methods on both evaluation metrics by a significant margin, demonstrating superiority in both generation fidelity and diversity. One possible explanation is that conventional intermediate representations of scene graphs used in previous methods, as exemplified by scene layouts, do not specifically optimize the semantic alignment between scene graphs and images. 
%
By contrast, SGDiff directly learns scene graph embeddings, maximizing both local and global semantic compliance with images in a self-supervised manner. Since our latent diffusion model incorporates these optimized embeddings to guide the generation process, we naturally achieve improved quality in image generation. We verify this hypothesis with rigorous ablation studies in \cref{sec:analysis}. We also test the two methods for conditioning on scene graph embeddings, and find that concatentation with the noisy latent code $z_t$ consistently outperforms concatenation with time embedding $t$.

\begin{table}[t]
  \setlength{\tabcolsep}{2.2pt}
  \small
  \begin{center}
  \begin{tabular}{l|cc|cc}
    \toprule 
    \multirow{2}{*}{Method} & \multicolumn{2}{c|}{\textbf{Inception Score} $\uparrow$}& \multicolumn{2}{c}{\textbf{FID} $\downarrow$} \\ 
     & COCO & VG & COCO & VG \\
    \midrule 
    Real Img (64$\times$64) & 16.3 $\pm$ 0.4 & 13.9 $\pm$ 0.5 & - & -\\
    Sg2Im \cite{johnson2018image} & 6.7 $\pm$ 0.1 & 5.5 $\pm$ 0.1 & 82.8 & 71.3 \\
    WSGC \cite{herzig2020learning} & 5.6 $\pm$ 0.1 & 8.0 $\pm$ 1.1&91.3&45.3\\
    SOAP \cite{ashual2019specifying} &7.9 $\pm$ 0.2&-& 65.3&-\\
    PasteGAN \cite{li2019pastegan} & 9.1 $\pm$ 0.2 & 6.9 $\pm$ 0.2 & 50.9 & 58.5 \\
      \arrayrulecolor{black!30}\midrule

    \textbf{SGDiff} (with $t$) &\textbf{10.6 $\pm$ 0.4}& \textbf{8.9 $\pm$ 0.5} &\textbf{26.8}&\textbf{27.5}\\
    \textbf{SGDiff} (with $z_t$) &\textbf{11.4 $\pm$ 0.4}& \textbf{9.3 $\pm$ 0.2} &\textbf{22.4}&\textbf{16.6}\\
      \arrayrulecolor{black}\midrule
    Real Img (128$\times$128) & 24.2 $\pm$ 0.9 & 17.4 $\pm$ 1.1 & - & -\\
    Sg2Im \cite{johnson2018image} & 7.1 $\pm$ 0.2 & 6.1 $\pm$ 0.1 & 93.3 & 82.7 \\
    WSGC \cite{herzig2020learning} & 5.1 $\pm$ 0.3 & 7.2 $\pm$ 0.3 & 108.6 & 80.4 \\
    SOAP \cite{ashual2019specifying} &10.4 $\pm$ 0.4&-& 75.4&-\\  
    PasteGAN \cite{li2019pastegan} & 11.1 $\pm$ 0.7&7.6 $\pm$ 0.7 & 70.7& 61.2  \\
      \arrayrulecolor{black!30}\midrule

    \textbf{SGDiff} (with $t$) &\textbf{13.1 $\pm$ 0.4}& \textbf{9.5 $\pm$ 0.5} &\textbf{32.7}&\textbf{29.6}\\
    \textbf{SGDiff} (with $z_t$) &\textbf{14.6 $\pm$ 0.9}& \textbf{11.4 $\pm$ 0.5} &\textbf{30.2}&\textbf{20.1}\\
      \arrayrulecolor{black}\midrule
    Real Img (256$\times$256) & 30.7 $\pm$ 1.2 & 27.3 $\pm$ 1.6 & - & -\\
    Sg2Im \cite{johnson2018image} & 8.2 $\pm$ 0.2 & 7.9 $\pm$ 0.1 & 99.1 & 90.5 \\
    WSGC \cite{herzig2020learning} & 6.5 $\pm$ 0.3 & 9.8 $\pm$ 0.4 & 121.7 & 84.1 \\
    SOAP \cite{ashual2019specifying} &14.5 $\pm$ 0.7&-& 81.0&-\\
    PasteGAN \cite{li2019pastegan} & 12.3 $\pm$ 1.0 &8.1 $\pm$ 0.9& 79.1 & 66.5  \\
      \arrayrulecolor{black!30}\midrule

    \textbf{SGDiff} (with $t$) &\textbf{16.0 $\pm$ 0.9}& \textbf{13.6 $\pm$ 0.7} &\textbf{40.1}&\textbf{36.4}\\
    \textbf{SGDiff} (with $z_t$) &\textbf{17.8 $\pm$ 0.8}& \textbf{16.4 $\pm$ 0.3} &\textbf{36.2}&\textbf{26.0}\\
      \arrayrulecolor{black}\bottomrule
  \end{tabular}
  \end{center}
  \savespace
  \caption{\textbf{Inception Scores and FIDs on COCO-Stuff and VG datasets.} Results of previous methods are either directly taken from their original papers or obtained by running official open-source implementations.}
  \label{exp-main}
\end{table}
 
\subsection{Qualitative Evaluations}
To explore SGDiff's ability of generating images that contain multiple objects and complex relations, we visualize some typical examples (with resolution 128$\times$128) in \cref{exp-qualitative}, where images are placed in ascending order of scene graph complexity. We compare SGDiff with a classical algorithm Sg2Im \cite{johnson2018image} and the state-of-the-art method PasteGAN \cite{li2019pastegan}. It is clear that images synthesized by SGDiff are not only more realistic, but also comply with the corresponding scene graphs better than both Sg2Im and PasteGAN. In contrast, images generated by Sg2Im are fuzzy and lack fine-grained details. Although PasteGAN can generate images with more details compared to Sg2Im, it tends to miss some important relations specified by the scene graphs, leading to obfuscated results.
Both methods are prone to generating images with wrong objects or relations. 
In contrast, our SGDiff consistently generates realistic images that match scene graphs well, since we rely on scene graph embeddings that are explicitly optimized for local and global semantic alignment. The generated objects have clear object boundaries and more visual details. We place more image samples in \cref{more-synthesis-detail}.

\begin{figure*}[ht]
    \centering
    \includegraphics[width=0.99\textwidth]{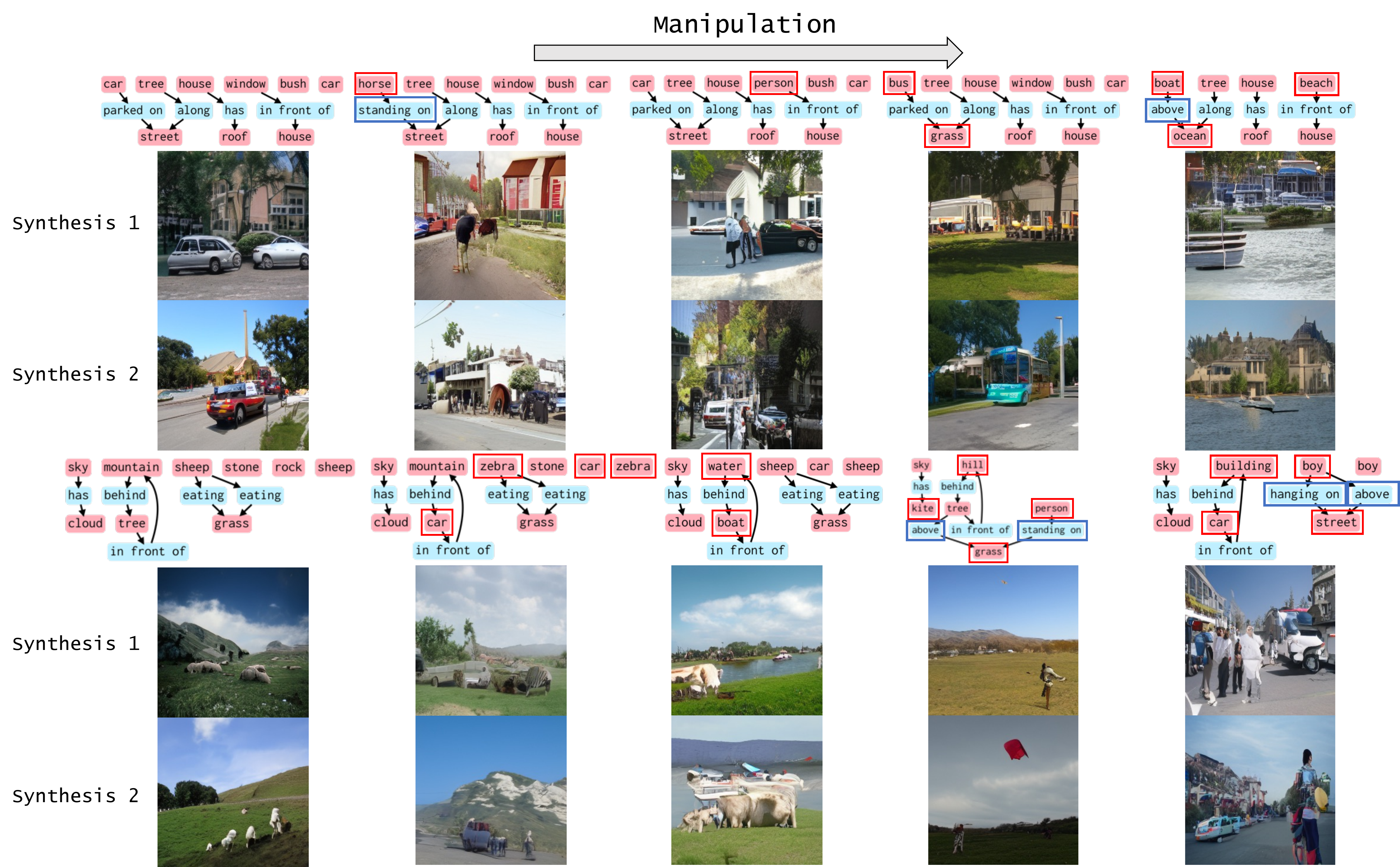}

    \caption{Semantic image manipulation (256$\times$256) with SGDiff. \textcolor{red}{Red} and \textcolor{blue}{blue} boxes denote object and relation modifications respectively.}
    \label{exp-manipulation}
\end{figure*}

\savespace
\paragraph{Semantic Image Manipulation.} To demonstrate the semantic consistency between generated images and scene graphs, we apply our SGDiff to manipulate image samples by modifying objects and relations in scene graph inputs. As shown in \cref{exp-manipulation}, SGDiff can not only produce compliant manipulation results (of resolution 256$\times$256) with respect to objects and relations, but also synthesize perceptually diverse images when conditioned on the same scene graph. The results demonstrate that our model can effectively leverage the scene graph embeddings learned through masked contrastive pre-training.

\subsection{Ablation Studies}\label{sec:analysis}
Key to our approach is learning scene graph embeddings via masked contrastive pre-training. Here we perform ablation studies to understand the importance of masked autoencoding loss and contrastive loss in training our scene graph encoder. We also empirically verify that our scene graph embeddings outperform manually crafted scene layout representations when combined with the same latent diffusion model on scene graph to image generation.
\paragraph{Retrieval Tasks.}
We first evaluate the impacts of masked autoencoding loss and contrastive loss on cross-modal semantic alignment through graph-to-image and image-to-graph retrieval experiments. In graph-to-image retrieval, we use scene graph embeddings and image embeddings to search the most semantically similar image for a given scene graph, then report the accuracy of finding the correctly paired image from the given dataset. The image-to-graph retrieval task is defined analogously. All results are provided in \cref{exp-ablation1}. 
Here ``obj.'' stands for experiments where we only train object embeddings, whereas ``Obj. + Rel.'' represents settings where we train both object and relation embeddings. We observe that using both embeddings boost the performance on all retrieval tasks. With only the contrastive loss, we can already obtain over 70\% accuracy in graph-to-image and image-to-graph retrieval. Adding masked auto-encoding loss further improves the accuracy, demonstrating better graph-image alignment.
\begin{table}[ht]
  \setlength{\tabcolsep}{1.5pt}

  \small
  \begin{center}
  \begin{tabular}{l|cc}
    \toprule 
    Average Accuracy 
     & Graph-to-Image & Image-to-Graph\\
    \midrule 
    Obj.&68.6&69.8\\
    Obj. + Rel.&70.3 (\textbf{+1.7})&70.6 (\textbf{+0.8})\\
    \arrayrulecolor{black!30}\midrule
    Contrastive &70.3 &70.6\\
    Contrastive + Masked&73.4 (\textbf{+3.1})&74.1 (\textbf{+3.5})\\
    \bottomrule
  \end{tabular}
  \end{center}
  \savespace
  \caption{Ablation study of masked contrastive pre-training on retrieval tasks.}
 
  \label{exp-ablation1}
\end{table}

\savespace
\paragraph{Generation from Scene Graphs.}
To understand the role of masked contrastive pre-training in scene graph to image generation, we train the same latent diffusion model with different scene graph embeddings: na\"{i}ve one-hot embeddings, layout embeddings, masked embeddings, constrastive embeddings, and combination of the last two. As for one-hot embeddings, we fix object and relation embeddings as one-hot vectors of their categories, then concatenate these embeddings together to condition the latent diffusion model. For conventional scene layout embeddings, we follow the same settings in PasteGAN \cite{li2019pastegan}. In \cref{tab:condition}, we report IS and FID scores for image samples of resolution 256$\times$256. We observe that embeddings obtained from either masked pretraining or contrastive pretraining outperforms one-hot embeddings or scene layout representations by a significant margin, and combining them together further improves sample quality.
\begin{table}[ht]    
  \setlength{\tabcolsep}{3pt}
  \small
  \begin{center}
  \begin{tabular}{l|cc}
    \toprule 
    Embeddings 
     & Inception Score $\uparrow$& FID $\downarrow$\\
    \midrule 
    One-Hot& 10.1 $\pm$ 1.3&87.1\\
    Layout& 12.3 $\pm$ 1.0& 52.7\\
    \midrule 
    Masked (Ours)&15.8 $\pm$ 0.6&26.2\\
    Contrastive (Ours)&16.1 $\pm$ 0.6&26.9\\
    Masked + Contrastive (Ours)&\textbf{16.4 $\pm$ 0.6}&\textbf{26.0}\\
    \bottomrule
  \end{tabular}
  \end{center}
 \vspace{-3mm}
 \caption{Ablation study of masked contrastive pre-training on scene graph to image generation.}
  \label{tab:condition}
\end{table}

\begin{figure}[ht]
    \centering
    \includegraphics[width=0.44\textwidth]{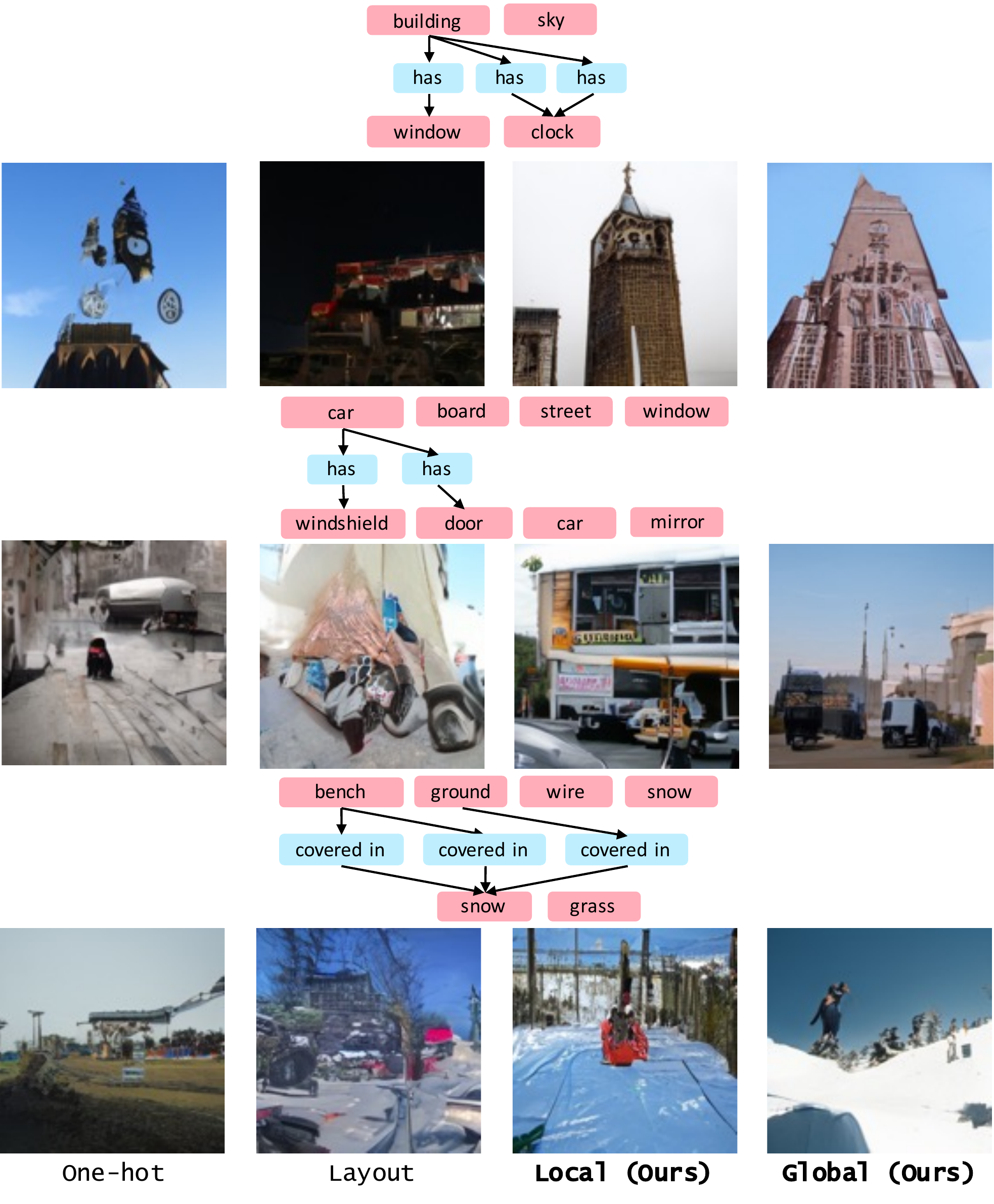}
    \vspace{-2mm}
    \caption{Latent diffusion with different SG embeddings.}
    \vspace{-3mm}
    \label{exp-condition}
\end{figure}

For qualitative comparison, we provide image samples with different scene graph embeddings in \cref{exp-condition}. Compared with one-hot embeddings and scene layout representations, we observe that embeddings from masked pre-training and contrastive pre-training enable SGDiff to generate images that are more realistic and comply better with the scene graphs. 
Compared to contrastive pretraining, we observe that embeddings obtained from masked pretraining tend to produce images that capture local structures better with more fine-grained object details, whereas embeddings from contrastive pretraining tend to focus more on matching the global structures at the overall image level.


\section{Conclusion}

This paper proposes a new framework, SGDiff, for image generation from scene graphs. SGDiff uses a masked contrastive pre-training approach to obtain scene graph emebeddings that allow for improved alignment between scene graphs and images, while also leveraging latent diffusion for improved scalability and generation quality. As a result, SGDiff produces more realistic and compliant images than previous methods that rely on manually crafted scene graph representations, such as scene layouts. SGDiff also makes image generation semantically controllable, allowing for easier manipulation of images through scene graph editing. Evaluation on standard datasets such as Visual Genome and COCO-Stuff show that SGDiff outperforms state-of-the-art approaches both qualitatively and quantitatively. 

{\small
\bibliographystyle{ieee_fullname}
\bibliography{main}
}
\onecolumn

\appendix

\vspace{-2cm}
\begin{center}
\LARGE\textbf{Supplementary Material}
\end{center}
\section{Implementation Details}
\label{imple-detail}
As illustrated in main text, our proposed framework SGDiff consists of three (pre-)training stages, masked contrastive pre-training for SG encoder, variational autoencoder pre-training for latent embedding of images, and latent diffusion training for SG-based image generation. 
Here we introduce the concrete network and optimization details of these stages in \cref{impl-details1} and \cref{impl-details2}. We use the same settings on both VG and COCO-Stuff datasets.

In masked contrastive pre-training, the nodes and edges of SGs are all preprocessed into 512-dimensional vectors for SG encoder. The ViT model tokenizes input images with patch size of 32$\times$32. Specifically, we set the ratio of random mask to 0.3 in masked autoencoding branch, and set the ratio between masked autoencoding loss and contrastive loss to 10:1 for facilitating the optimization.
In variational autoencoder pre-training, we embed input images into compact latents with a downsampling factor of 8, and maximize the decoding ability by optimizing the MSE objective. And the ratio between KL divergence and MSE is set to 8:10.
In latent diffusion training, we use cross-attention mechanism for conditional diffusion process in all experiments. 
\begin{table}[ht]
  \setlength{\tabcolsep}{3pt}
  \begin{center}
  \begin{tabular}{lllccccc}
    \toprule 
    Stage & Module
     & Backbone & Input & Output&Blocks \\
    \midrule 
     \multirow{2}{*}{Masked Contrastive Pre-Training}&Image Encoder&ViT-B/32&256$\times$256$\times$3&512&11\\&SG Encoder& Graph Convolution&512&512&5\\
     \midrule
     \multirow{2}{*}{Variational Autoencoder Pre-Training}&Encoder&ResNet&256$\times$256$\times$3&32$\times$32$\times$4&3\\&Decoder&ResNet&32$\times$32$\times$4&256$\times$256$\times$3&3\\
     \midrule
     \multirow{3}{*}{Latent Diffusion Training}&UNet-Encoder&ResNet + CrossAttention&32$\times$32$\times$4&8$\times$8$\times$1024&9\\&UNet-MiddleBlock&ResNet&8$\times$8$\times$1024&8$\times$8$\times$1024&3\\&UNet-Decoder&ResNet + CrossAttention&8$\times$8$\times$1024&32$\times$32$\times$4&10\\
    \bottomrule
  \end{tabular}
  \end{center}
  \savespace
  \caption{Network Details of SGDiff.}
 
  \label{impl-details1}
\end{table}

\begin{table}[ht]
  \setlength{\tabcolsep}{3pt}
  \begin{center}
  \begin{tabular}{lccccc}
    \toprule 
    Stage & Optimizer&	Batch Size&	LR	&	Loss \\
    \midrule 
     Masked Contrastive Pre-Training&	Adam	&16&	5.0 e$^{-4}$&Masked Autoencoding : Contrastive = 10 : 1\\
    Variational Autoencoder Pre-Training&Adam	&128	&1.0 e$^{-5}$&KL Divergence : MSE = 8 : 10\\
    Latent Diffusion Training&Adam&	16	&1.0 e$^{-6}$& $L_2$ Distance\\
    \bottomrule
  \end{tabular}
  \end{center}
  \savespace
  \caption{Optimization Details of SGDiff.}
 
  \label{impl-details2}
\end{table}
\section{More Synthesis Results} 
\label{more-synthesis-detail}
\begin{figure*}[t]
    \centering
    \includegraphics[width=0.95\textwidth]{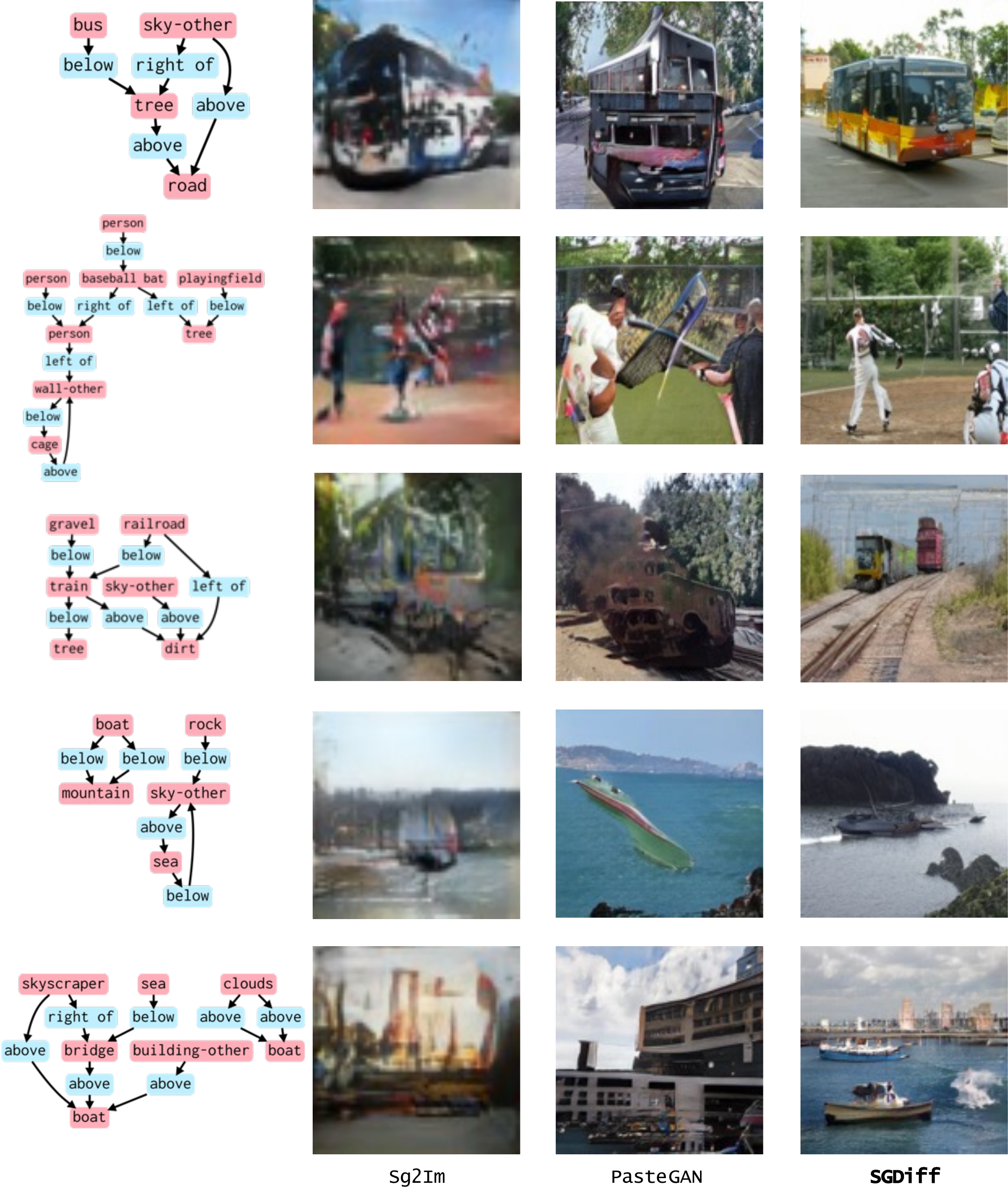}
    \caption{Image samples (256$\times$256) generated by Sg2Im \cite{johnson2018image}, PasteGAN \cite{li2019pastegan}, and our SGDiff given the same scene graphs.}
    \label{coco1}
    \vskip -0.15in
\end{figure*}

\begin{figure*}[t]
    \centering
    \includegraphics[width=0.95\textwidth]{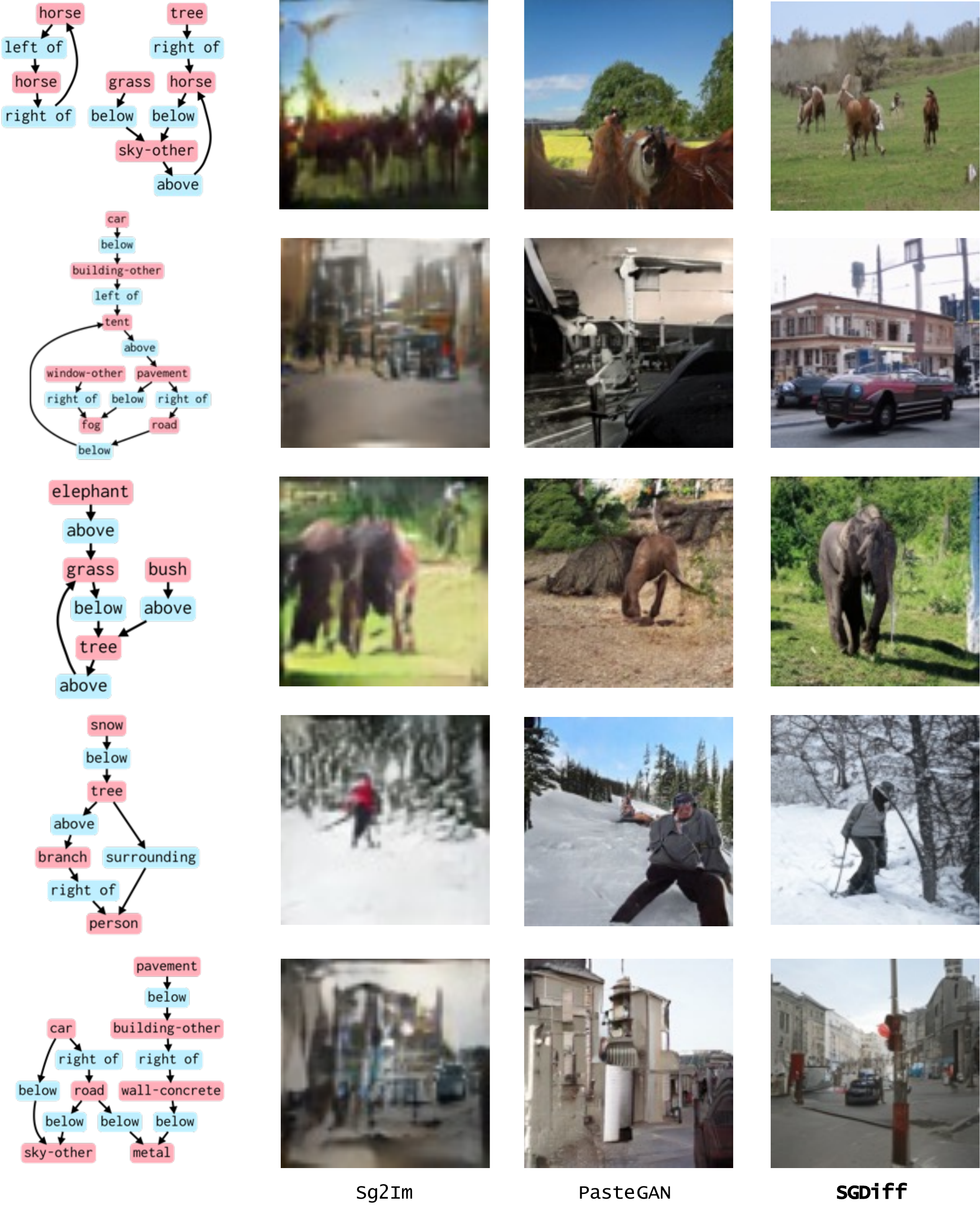}
    \caption{Image samples (256$\times$256) generated by Sg2Im \cite{johnson2018image}, PasteGAN \cite{li2019pastegan}, and our SGDiff given the same scene graphs.}
    \label{coco2}
    \vskip -0.15in
\end{figure*}

In previous qualitative evaluations, we have shown the synthesis comparison results on VG dataset. Here, we provide more results on COCO-Stuff dataset with the resolution of 256$\times$256 in \cref{coco1} and \cref{coco2}.  
SGDiff exhibits the superiority of generation quality on COCO-Stuff dataset, and SGDiff can generate images that are more realistic and semantic-compliant than previous methods Sg2Im \cite{johnson2018image} and PasteGAN \cite{li2019pastegan}. The results not only reveal the SG embeddings learned by our masked contrastive pre-training are effective in graph-image semantic alignment, but also demonstrate the efficacy of our local-global conditional latent diffusion.  

\end{document}